# Non-anchor-based vehicle detection for traffic surveillance using bounding ellipses

Byeonghyeop Yu, Johyun Shin, Gyeongjun Kim, Seungbin Roh and Keemin Sohn

*Abstract*—Cameras for traffic surveillance are usually pole-mounted and produce images that reflect a birds-eye view. Vehicles in such images, in general, assume an ellipse form. A bounding box for the vehicles usually includes a large empty space when the vehicle orientation is not parallel to the edges of the box. To circumvent this problem, the present study applied bounding ellipses to a non-anchor-based, single-shot detection model (CenterNet). Since this model does not depend on anchor boxes, non-max suppression (NMS) that requires computing the intersection over union (IOU) between predicted bounding boxes is unnecessary for inference. The SpotNet that extends the CenterNet model by adding a segmentation head was also tested with bounding ellipses. Two other anchor-based, single-shot detection models (YOLO4 and SSD) were chosen as references for comparison. The model performance was compared based on a local dataset that was doubly annotated with bounding boxes and ellipses. As a result, the performance of the two models with bounding ellipses exceeded that of the reference models with bounding boxes. When the backbone of the ellipse models was pretrained on an open dataset (UA-DETRAC), the performance was further enhanced. The data augmentation schemes developed for YOLO4 also improved the performance of the proposed models. As a result, the best mAP score of a CenterNet exceeds 0.95 when augmenting heatmaps with bounding ellipses.

## I. Introduction

The success of traffic operation and maintenance totally depends on whether area-wide traffic surveillance can be secured without failure. Much effort has been invested in space-based approaches to obtain such a level of integrity for traffic surveillance. In this context, the most conventional way to measure traffic volumes and speeds utilizes an inductive loop detector embedded in the road surface [1]-[5]. This approach, however, does not guarantee error-free detection and entails large maintenance costs. There are special detectors that utilize an ultrasonic beam [6],[7] and piezo-electronic sensors [8],[9] that can overcome the handicap of loop detectors, but these are less cost-efficient and have failed to achieve an acceptable level of market penetration. Meanwhile, traffic surveillance methods are converging on the use of computer-vision schemes, as deep-learning technologies have improved the performance of vehicle detection.

Computer-vision technology had been widely adopted in traffic surveillance even before recent advances in deep-learning models. The most popular technology prior to deep learning has been the background subtraction method, wherein vehicles are recognized from a silhouette image created by subtracting pixel values of an image of concern from those of a background image [10]-[12]. The success of the method depends on the accurate creation of a background image, and the most prevalent way to accomplish that has been to synthesize a virtual image using consecutive video frames taken at the same site. Each pixel adopts the average (or mode) of the corresponding pixel values for the video frames. A virtual background image, however, cannot be constant due to varying conditions of lighting, weather, and camera positions and angles. There have been many miscellaneous attempts to overcome these problems [10]-[15], but no ultimate solution has been found. The optical-flow method is another popular computer-vision technology that is used to detect vehicles and their speeds [16],[17]. This method derives gradients of pixel values to trace the movement of vehicles between consecutive video frames in a mathematical manner. The optical-flow method had been relatively successful in recognizing moving vehicles, and, thus, in measuring traffic parameters, but its popularity has been diminished by the emergence of deep-learning technologies.

In the early 2010s there was a great leap in object detection owing to the advance of deep-learning technologies. The vehicle detecting model for traffic surveillance is rapidly converging to the adoption of deep-learning technologies [18]-[22]. Details of this trend will be described in the next section with reviews of related work. The present study suggests a novel methodology to improve the performance of a deep-learning model that detects vehicles from video images taken for traffic surveillance. The key to this novel methodology is to represent vehicle images with bounding ellipses rather than conventional bounding boxes. It should be noted that this measure might not work as a general object detector that recognizes objects of various shapes. In the present study, we focused only on detecting vehicles for traffic surveillance,

This research was supported by the Chung-Ang University Research Scholarship Grants in 2019 and in part by the Korea Agency for Infrastructure Technology Advancement (KAIA) grant funded by the Ministry of Land, Infrastructure and Transport (Grant 20TLRP-B148676-03). *(Corresponding author: Keemin Sohn.)*

The authors are with the Laboratory of Big-data Applications in Public Sector, Chung-Ang University, Seoul 06974, South Korea (e-mail: ybh22@cau.ac.kr; olfy1021@cau.ac.kr;kjunn7033@cau.ac.kr;sbr444@cau.ac.kr; kmsohn@cau.ac.kr).

wherein vehicles take the shape of an ellipse since every camera for traffic surveillance is poll-mounted and provides images from an aerial view. Video frames for traffic surveillance include moving vehicles, and the moving direction of vehicles generally does not match either edge of the video frames. For this reason, bounding boxes are likely to include large empty spaces that are not occupied by vehicles.

Recently, an oriented bounding box (OBB) has been adopted in localizing objects in aerial photos [23]-[25]. There are two ways to represent an OBB. The angle of objects from a baseline is added to the width and height when embedding a regression head. On the other hand, the 8 coordinates of the 4 vertices of a quadrilateral can be directly fitted to the ground-truth values. Basically, the utility of OBBs equals that of the bounding ellipses adopted in the present study. The results of the present study enhanced the utility of bounding ellipses in detecting vehicles for traffic surveillance by incorporating several technologies such as a non-anchor-based detector, a heatmap augmented with bounding ellipses, and "state-of-the-art" data augmentation schemes.

One of the obstacles to using an OBB involves computing an intersection over union (IOU) between OBBs, which requires more time than adopting horizontal bounding boxes. How to deal with the discontinuity of the rotation angle of vehicles is another key to successfully utilizing the OBB. Adopting a non-anchor-based detector in the present study was a solution for the former problem. A CenterNet detection model [28] requires neither anchor boxes nor the non-max suppression (NMS) process for inference. The NMS process includes many repeated computations of IOUs. It thus would entail long computation times when used together with bounding ellipses (or OBBs). The latter problem was resolved by other researchers who changed the rotation angle regression into a classification task, albeit at the expense of losing the angular granularity [26]. Chen et al. [27] proposed a more fundamental solution to overcome the discontinuity in rotation angle and the difficulty in analytically computing an IOU. They devised a pixels-IOU (PIOU) loss that could be minimized to derive the size and orientation of objects instead of using a regression loss that directly fits them, wherein an IOU between two OBBs is computed on a pixel-by-pixel basis. This scheme also requires more computing time due to the pixel-wise computation of IOUs.

The proposed approach makes a significant contribution by modifying the way a heatmap is used with CenterNet. The ground-truth heatmap was drawn according to various sizes and orientations of vehicles. A multi-variate Gaussian density function was used to fill bounding ellipses in a heatmap. Since labeling heatmaps with bounding ellipses is a preprocess for training, it never increases the inference time for vehicle detection. Basically, using bounding ellipses is identical to employing OBBs in that both utilize the regression of an object's orientation angle in addition to the width and height. However, the proposed approach differs from the existing object-detection models based on OBBs in that the graphically labeled heatmaps filled with bounding ellipses would assist the regression to fit the width, height, and orientation of the objects in question. We made heatmap information compatible with the objective of a regression head. It would be inefficient if a heatmap provided only the center point of an object and a regression head attempted to fit the width and length of the oriented object as well as the orientation itself. A SpotNet that incorporates an additional segmentation head to CenterNet was also tested with bounding ellipses. The SpotNet model was expected to increase the effect of the augmented heatmap in detecting vehicles by adding a segmentation loss for the full-size image.

We also applied a combination of 'state-of-the-art' data augmentation schemes to training the proposed models and reference models. This routine task resulted in the best combination of data augmentation schemes to enhance the vehicle detection performance. Although we do not present a newly designed neural network model, our contributions to enhancing the vehicle detection accuracy for traffic surveillance are four-fold as follows.

- Non-anchor-based detector models (CenterNet and SpotNet) are adapted to accommodate bounding ellipses without increasing the computing time for inference.
- The vehicle detection performance of CenterNet is enhanced by graphically labeling heatmaps with bounding ellipses.
- The most desirable combination of data augmentation methods was found for vehicle detection for the purpose of traffic surveillance.
- As a result, the vehicle detection performance, when measured by the mean average precision (mAP), was increases to an exceptional level (=0.953).

Two other "state-of-the-art" object detection models (YOLO4 and SSD) were mobilized as references to confirm the superiority of the proposed approach. The PIOU loss [27], which was devised to circumvent the difficulty in dealing with oriented bounding boxes, was also tested using CenterNet when heatmaps augmented with bounding ellipses were employed as the ground truth.

The next section introduces cutting-edge technologies for deep-learning-based object detection. How "state-of-the-art" methods have been applied to traffic surveillance is also explained in the next section. The third section describes the architectures of CenterNet and SpotNet detection models and accounts for how the models should be revised to accommodate bounding ellipses. The fourth section explains how data are prepared and labeled for training and testing models. Several methods of data augmentation are introduced in the same section. The fifth section compares results based on mAP scores that has been adopted as a globally accepted performance measure for object detection. The last section draws overall conclusions and provides suggestions for further studies to enhance the detection performance.

## II. RELATED WORK

The mainstream approach to traffic surveillance is rapidly converging to deep-learning-based computer vision technologies [18],[20]-[22]. In computer vision studies, detecting objects in an image has long been regarded as a difficult task. The performance of deep-learning-based object detection models has, however, already surpassed human ability owing to the ever-growing advancement in deep-learning technologies. The ″state-of-the-art″ detection algorithm can be categorized into two groups. Models that belong to the first category separate the region proposal task from the subsequent classification module. In the initial stage of developing these types of models, all potential regions in an image that might include an object are determined using a rule-based manner, and a learning model classifies objects for the proposed regions [29]. Both tasks were integrated to a single framework later, and region-proposal models are also trained on data to distinguish the foreground from the background [30],[31]. A Faster-RCNN is a two-stage model with region-proposals that has shown the best detection performance but is handicapped by a relatively long computation time for inference.

The second category includes one-stage detection models developed to speed up the inference time at the expense of deteriorating detection accuracy. One-stage detectors simultaneously conduct both the localization and classification tasks in an end-to-end manner. The YOLO series is the most popular form of the one-stage model [32]-[34]. A YOLO model reduces the detection time by using a grid to divide the input image and assigning several anchor boxes to each grid for detection. An anchor box is a predefined bounding box, and its location and shape are adjusted during learning. Early versions of YOLO did not outperform the two-stage model in accuracy, but the latest version (YOLO4) has recorded equivalent, or even better, performances by reinforcing the model architecture and adopting diverse data augmentation schemes for training [35].

A single-shot multi-box detector (SSD) is another successful version of the one-stage model [36]. The SSD also depends on anchor boxes to detect objects. The difference from a YOLO model is that a SSD can separately detect objects in an image by different scales. That is, several intermediate feature maps chosen from a deep neural network pipeline, each of which has a different resolution, are used to separately detect objects of different sizes, whereas a YOLO model uses only the last feature map for detection. This is why the title includes the word "multi-box". A RetinaNet constitutes another axis of the one-stage detection models [37]. RetinaNet first adopted a focal-loss approach to reduce the risk of over fitting by assigning different weights to each pixel according to the presence or absence of an object.

All the one- and two-stage models introduced above depend on anchor boxes, which can create complexity due to a large number of anchor boxes. The use of anchor boxes is also accompanied by the burden of determining many hyper-parameters such as the number, size, and shape of anchor boxes. Some researchers have developed a one-stage detection model that is free from anchor boxes. The CornerNet model uses a novel concept of key-points without the need to employ anchor boxes [38]. With this approach, the two corner points of a bounding box are directly predicted based on focal losses. This scheme removes the necessity of an NMS process for inference. The CornerNet model, however, requires the additional task of matching the upper-left corners to their corresponding lower-right corners to constitute bounding boxes for detecting objects, which entails an exhaustive amount of computation time. Some researchers overcame the complication by developing a robust key-point-based object detector (CenterNet) [28]. A CenterNet uses only a single key-point (=center point) to recognize an object. There are two distinct advantages for the model compared with the existing one-stage detectors. First, the CenterNet does not use anchor boxes. Second, there is no need to implement an NMS for the final inference, which repeatedly computes the IOU between the estimated and observed bounding boxes. As an extension of a CenterNet, some researchers developed a SpotNet model by adding a head for semantic segmentation to its architecture [39]. When training the model, the head for semantic segmentation is fed with a silhouette derived by a background subtraction method using consecutive video shoots.

Among the two-stage detectors introduced above, a Faster-RCNN recorded the best performance and has been used as a vehicle detector for traffic surveillance [40]-[42]. As a detector, however, the RCNN requires computation time so large that it cannot be used in real-time applications. Measuring vehicle speeds and traffic volumes in the field requires the tracking of each vehicle in a very short interval. Thus, the YOLO series has been widely adopted in studies of traffic surveillance [43]-[47]. The performance of YOLO models ranges between 0.6 and 0.8 when measured by the mAP score. CenterNet and its extension SpotNet have also been adopted in studies of vehicle detection using the UA-DTRAC dataset [20],[39],[48]-[49]. These two key-point-based detection models have recorded mAP scores exceeding 0.8, which is higher than those for any other detector.

In the present study, we chose the latter two detection approaches (CenterNet and SpotNet) to confirm the advantage of bounding ellipses, since they require neither the computation of IOUs nor a regression of the corner points of bounding boxes. The focal loss of the RetinaNet approach was applied to original heatmaps and those augmented with bounding ellipses. As a baseline, the two models were also trained and tested with bounding boxes. The remaining one-stage detection approaches (YOLO4 and SSD) with bounding boxes were also used as references to validate the performance of the two models that used bounding ellipses.

### III. MODIFYING THE MODELLING FRAMEWORK OF CENTERNET

In the present study, we adopted a key-point-based detection approach (CenterNet) to verify the utility of replacing bounding boxes with ellipses. In this section, the architecture and loss function of the CenterNet approach is modified for using bounding ellipses. CenterNet can employ several different backbones such as ResNet-18, ResNet-101, DLA-34, and Hourglass-104. The present study adopted an Hourglass-104 because in an earlier study [23] it outperformed other backbones based on mAP scores. The original Hourglass-104 is, however, too heavy to be applied to detecting vehicles for online traffic surveillance. A lighter backbone was set up by reducing the depth and width of the original Hourglass-104 as shown in Fig. 1. The number of parameters of the new backbone is downsized to 36,588,000, whereas that of the original one was tantamount to 188,409,000. A preliminary experiment revealed that the mAP score is improved from 0.733 to 0.853 even though the reduced backbone was adopted.

The head of CenterNet detection is composed of a heatmap and an embedding for regression, each of which minimizes a different loss function. The specification of each loss function will be addressed later in this section. Fig.2 shows the model architecture used in the present study. The model architecture becomes a SpotNet if the shaded region of Fig. 2 is included. A semantic segmentation head was added to the two exiting heads of the CenterNet model under the expectation that the full-size segmentation information would reinforce the role of heatmap. Unlike the original SpotNet that acquired a background image from consecutive video frames, in the present study bounding ellipses were directly used to fill the region for vehicle presence.

The input for CenterNet is a 3-channel color image with a fixed size (W×H×3). W is the width and H is the height of the input image as measured by the number of pixels. A batch of input images, of course, can be used when training and inferring a model. The first head of the CenterNet is a center-point heatmap obtained after layer-by-layer convolutions are applied to an input image for encoding and decoding in the backbone. The dimensions of the output heatmap are reduced to (W/R×H/R×C) after passed through the backbone. R is the output stride and C denotes the number of categories. In the present study, three vehicle categories were selected: cars, buses and trucks. Consequently, after going through the backbone, the size of the input image was reduced to a smaller dimension that would be tractable in the subsequent detection process.

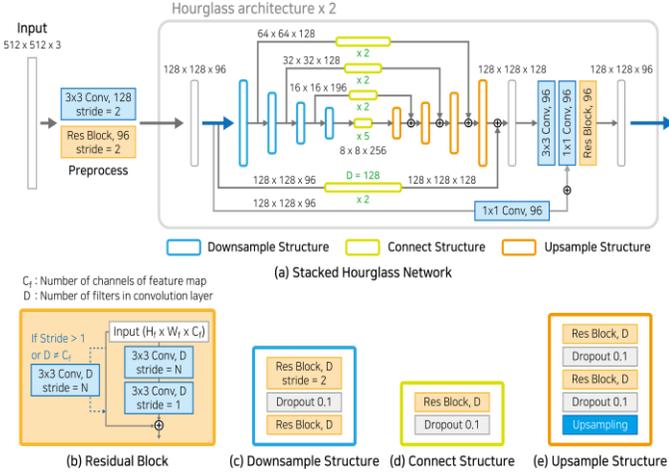

Fig. 1. The reduced architecture of Hourglass-104 for the use of the backbone of CenterNet

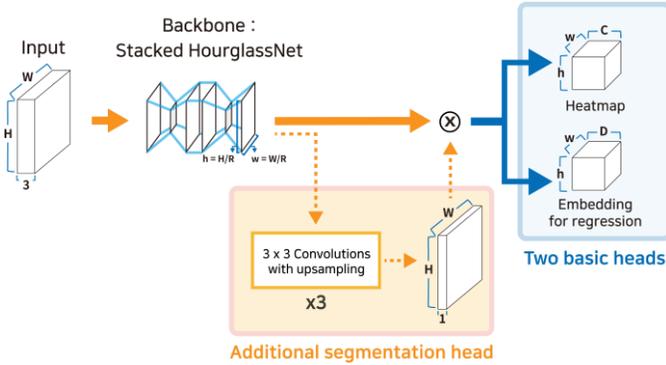

Fig. 2. The architecture of CenterNet

When assigning ground-truth information to heatmaps in the original CenterNet, cells around the center point of a bounding box showed an independent Gaussian density, which draws a circle regardless of the actual shape of vehicle. On the other hand, in the present study bounding ellipses were utilized to fill the pixels of heatmaps. The ground-truth heatmap was drawn so that the orientation and shape of the bounding ellipses could be reflected. Fig.3 displays three labels to represent a ground-truth bounding ellipse, which correspond to two lengths of the first and second axes and the orientation of the first axis.

In theory, drawing a ground-truth heatmap based on the three labels is simple. A multivariate Gaussian density function can fill internal cells of a bounding ellipse in the heatmap, which corresponds to a vehicle. Eq. (1) represents the Gaussian density and guarantees that a cell value that corresponds to a center point should have the maximum ground-truth value (=1). The intensity of pixels gradually diminishes as they move away from the center point and approach 0 when reaching the ellipse boundary. The maximum intensity could be adjusted when label smoothing [45] or CutMix augmentation [46] is applied to the training data.

$$y_{ijc} = e^{-\left[\left(\frac{\cos^2\theta}{2\sigma_{cx_k}^2}+\frac{\sin^2\theta}{2\sigma_{cy_k}^2}\right)(i-cx_k)^2+\left(\frac{\sin^2\theta}{2\sigma_{cx_k}^2}+\frac{\cos^2\theta}{2\sigma_{cy_k}^2}\right)(j-cy_k)^2 + 2\left(\frac{-\sin2\theta}{4\sigma_{cx_k}^2}+\frac{\sin2\theta}{4\sigma_{cy_k}^2}\right)(i-cx_k)(j-cy_k)\right]} \quad (1)$$

$$(cx_k, cy_k) = \left(\left\lfloor\frac{CX_k}{R}\right\rfloor, \left\lfloor\frac{CY_k}{R}\right\rfloor\right) \quad (2)$$

In Eq. (1), $y_{ijc}$ is the intensity of the $(i,j)$ pixel of a ground-truth heatmap for category $c$, $\theta$ is the orientation angle of a bounding ellipse, and $(cx_k, cy_k)$ are the center coordinates of a bounding ellipse in a heatmap and can be computed using Eq. (2) where $(CX_k, CY_k)$ are the center coordinates in an original image. $(\sigma_{cx_k}, \sigma_{cy_k})$ denotes the bandwidths of major and minor axes, respectively. The bandwidths are set as $(l_1/6R, l_2/6R)$, so that the pixel intensity approaches 0 at the ellipse boundary according to the dictates of a Gaussian density. A significant contribution of the present study is the creation of ground-truth heatmaps that are more consistent with three labels of abounding ellipses in the regression head. More concretely, this ellipse-driven heatmap eases the regression for predicting the width, height, and orientation of vehicles.

We devised a robust annotation tool that makes the drawing of bounding ellipses as simple as drawing bounding boxes. Fig. 4 intuitively shows the advantages of adopting bounding ellipses rather than bounding boxes. For most cases, a bounding box is less efficient than a bounding ellipse, since it may encompass large empty spaces. In addition, labels of the width and height of a bounding box cannot account for the actual size of a vehicle. On the other hand, a bounding ellipse can have labels that are consistent with the actual size of a vehicle.

Moreover, a false intersection area is generated when vehicles are bounded with boxes, even though right two vehicles in Fig. 4. do not overlap. No false intersections are generated when using bounding ellipses. These merits motivated us to replace bounding boxes with bounding ellipses. However, it should be noted that using bounding ellipses is effective only when adopting a key-point-based object detector wherein no IOU computation is required. Applying bounding ellipses to an anchor-based detector would increase the computation complexity.

A focal loss was applied to heatmaps for both center-point detection and classification, which was first devised in the RetinaNet [38]. Eq. (3) denotes the definition of focal loss used in the original CenterNet.

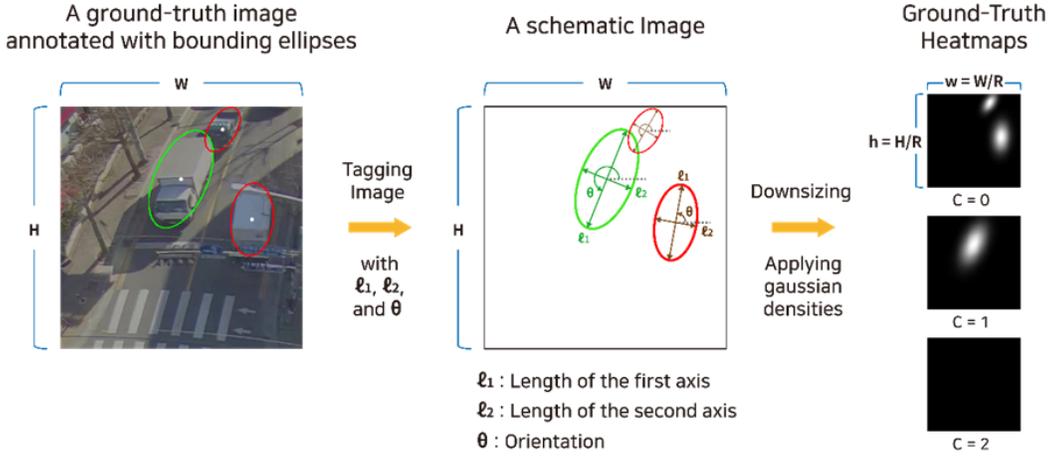

Fig. 3. Generating ground-truth heatmaps

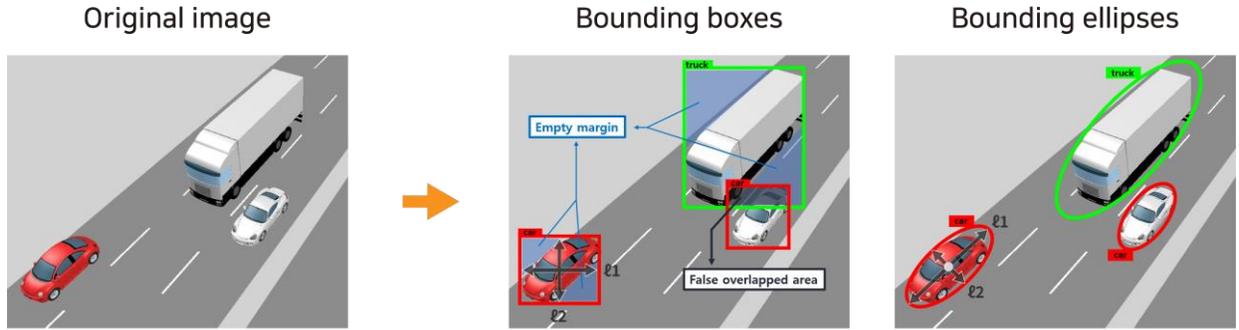

Fig. 4. The superiority of bounding ellipses to bounding boxes for representing vehicles.

$$\mathcal{L}_{focal} = \frac{-1}{N}\sum_c \sum_{ij} \begin{cases} (1-\hat{y}_{ijc})^\alpha \log(\hat{y}_{ijc}) & if\ y_{ijc} = 1 \\ (1-y_{ijc})^\beta (\hat{y}_{ijc})^\alpha \log(1-\hat{y}_{ijc}) & Otherwise \end{cases} \quad (3)$$

In Eq. 3, $\hat{y}_{ijc}$ and $y_{ijc}$ are the predicted and ground-truth cell values of a heatmap, respectively. $N$ is the number of bounding ellipses to be detected in a training batch of images. The focal loss is a variant of cross-entropy loss, wherein the presence and absence of vehicles are weighted differently in order to avert over fitting. For ground-truth center points, $(1-\hat{y}_{ijc})^\alpha$ assigns more weight to the loss of vehicle presence when the predicted intensity of a vehicle presence is low and grants less weight to the loss when the predicted intensity is high. For other points, $(\hat{y}_{ijc})^\alpha$ exaggerates the loss of vehicle absence when the predicted intensity of vehicle absence is low and diminishes the loss when the predicted intensity is high. $(1-y_{ijc})^\beta$ is the weight that is used to decrease the loss of vehicle absence for non-zero cells in ground-truth heatmaps, which belong to the region within a bounding ellipse. Hyper-parameters $\alpha$ and $\beta$ were set at 2 and 4, respectively, as in the original setup [38].

The other head of a CenterNet is an embedding for regression. The first regression loss is to adjust the positioning error of the center points due to the reduced size of the heatmap. This regression is inevitable because the center point is not predicted in an original image with the size of $(W \times H)$ but, rather, in a heatmap with the down-sampled size of $(w \times h)$ where $w = \frac{W}{R}$ and $h = \frac{H}{R}$. The predicted offset is used to adjust a predicted center point in a heatmap, and then the adjusted coordinates are magnified to reproduce the corresponding center position in the original input image. For this offset regression, only cells in a heatmap that correspond to center points is considered. Eq. (4) denotes the offset loss to be minimized while training a model. The embedding dimension of the offset regression should be $(w \times h \times 2)$ to accommodate differences both in horizontal and vertical coordinates.

$$\mathcal{L}_{offset} = \frac{-1}{N}\sum_k |\hat{o}_k - o_k| \quad (4)$$

In Eq. (4), $\hat{o}_k$ is the predicted offset for the center point of the $k^{th}$ bounding ellipse, $o_k$ is the target offset computed by $(\frac{CX_k}{R} - \lfloor\frac{CX_k}{R}\rfloor, \frac{CY_k}{R} - \lfloor\frac{CY_k}{R}\rfloor)$, and $|\ |$ denotes a smooth $L_1$ operator. At an inference time after training, a predicted center point for a heatmap is adjusted using the predicted offset $o_k$.

The second regression is intended to match the size and orientation of bounding ellipses. This scheme is another distinction of the present study from the original CenterNet. The original CenterNet matches the predicted width and height of bounding boxes to the ground truth. This choice is inefficient, however, as illustrated in Fig. 4, because the width and height of a bounding box are not directly associated with the size of a target object. To find these dimensions, a deep net first must recognize the edges of a bounding box that face the end portion of an object, and then infer the location and size of the box. The present study adopted a loss function to directly minimize the difference in the axis lengths between predicted and ground-truth bounding ellipses. The mathematical meaning of finding the two axes of a bounding ellipse in a heatmap is to derive both eigenvectors of the covariance matrix for cell points within the bounding ellipse. This computation is compatible with conducting the conventional principal component analysis (PCA), which can be easily accomplished using a neural network with a single hidden layer.

Two channels of embedding are necessary to accommodate both lengths of the two axes of a bounding ellipse. The orientation regression adds an additional channel, and, thus, the dimension of regression embedding becomes ($w \times h \times 3$). If the previous offset channels are integrated, a total of 5 channels ($D=5$) are necessary for the regression embedding. It should be noted that each regression loss is defined only for the center points of bounding ellipses. Eq. (5) denotes the size and orientation loss.

$$\mathcal{L}_{size\_ori} = \frac{-1}{N}\sum_k |\hat{s}_k - s_k| \quad (5)$$

In Eq. (5), $\hat{s}_k$ and $s_k$ are the predicted and ground-truth tensors with sizes of ($3 \times 1$), each of which represents the two axis lengths and the orientation of bounding ellipses.

As mentioned earlier, regression with the three labels of a bounding ellipse is basically identical to using an OBB, whereas computing an IOU between ellipses is more complex than computing that between OBBs. Some researchers have suggested a pixel-based IOU loss that can substitute the regression loss of the size and orientation of OBBs [27]. In the present study, as an alternative to the regression of the size and orientation of bounding ellipses, a PIOU loss was applied to a CenterNet for comparison. Because the PIOU loss is obtained from pixel-wise computations, computing times are onerous. Therefore, a kernel approximation was adopted to compute the PIOU loss. Such an approximation is, however, a cause of detection failure for small objects in terms of our experimental results. The PIOU loss has the advantage of circumventing the discontinuity problem raised when directly fitting an orientation angle. Eq. (6) denotes a PIOU loss function.

$$\mathcal{L}_{PIOU} = \frac{-1}{N}\sum_k \ln \frac{\cap_k}{\hat{s}_k[0]\hat{s}_k[1]+s_k[0]s_k[1]-\cap_k} \quad (6)$$

In Eq. (6), the denominator is the union area of the estimated and ground-truth OBBs, and the symbol ($\cap_k$) in the numerator indicates the intersection area of both. The intersection area is approximately computed by using a kernel function. For details of computing the PIOU loss, readers should refer to Chen et al. [27].

For a SpotNet, an extra head is added to accommodate the semantic segmentation. Whereas other heads are connected directly from the last feature map of the backbone network, the segmentation head up-samples the last feature map of backbone to generate an output feature map with the same size as the input image (see Fig. 2). The binary cross entropy loss is minimized for the semantic segmentation. Eq. (7) denotes the loss of segmentation.

$$\mathcal{L}_{seg} = \frac{-1}{W \times H}\sum_p [y_p \log(\hat{y}_p) + (1-y_p)\log(1-\hat{y}_p)] \quad (7)$$

In Eq. (7), $\hat{y}_p$ and $y_p$ represent the predicted and ground-truth cell values, respectively, for the $p^{th}$ position in an output segmentation map of an original size ($W \times H$).

Finally, the total loss, $\mathcal{L}_{tot}$, is set up in Eq. (8) by integrating all losses introduced above. The last term in the total loss is included only when SpotNet is used. In addition, $\mathcal{L}_{size\_ori}$ and $\lambda_{size\_ori}$ can be replaced with $\mathcal{L}_{PIOU}$ and $\lambda_{PIOU}$, respectively. The result of using the PIOU loss instead of the regression loss will be discussed in section V.

$$\mathcal{L}_{tot} = \mathcal{L}_{focal} + \lambda_{offset}\mathcal{L}_{offset} + \\ \lambda_{size_{ori}}\mathcal{L}_{size_{ori}}(\text{or }\lambda_{PIOU}\mathcal{L}_{PIOU}) + \mathcal{L}_{seg} \quad (8)$$

In Eq. (8), $\lambda_{offset}$ and $\lambda_{size\_ori}$ are the relative weights for offset and size losses set at 1.0 and 0.1, respectively, following the training scheme of the original CenterNet. $\lambda_{PIOU}$ is also set at 0.1 according to [27].

## IV. TESTBED AND DATA PREPARATION

Deep learning models for vehicle detection have an advantage whereby the number of vehicle types to be classified is relatively small. 3 vehicle types were considered in this study. Nonetheless, vehicle detection that can work everywhere does not exist at the current stage of traffic surveillance. Most previous studies have attempted to train their models on a site-by-site basis [41]-[43], and, at least, to fine-tune each model with local data after pretraining on an open dataset [38]. For a new labeling task with bounding ellipses, we devised a simple annotation tool. Thus, the labeling task with this tool would be as easy as the task of bounding boxes.

The present study begins with the difficulty in securing a universal vehicle detector for traffic surveillance at real sites located in Bucheon, South Korea. At the initial state of the project, we expected existing "state-of-the-art" object detectors to work well, because these had been trained on large-scale open datasets such as COCO [50] and PASCAL VOC [51]. We realized, however, that such detection models cannot be fully qualified without being fine-tuned using local images on a site-by-site basis. This is the "status quo" of deep learning technologies for traffic surveillance. In the same context, labeling with bounding ellipses were conducted for a specific site.

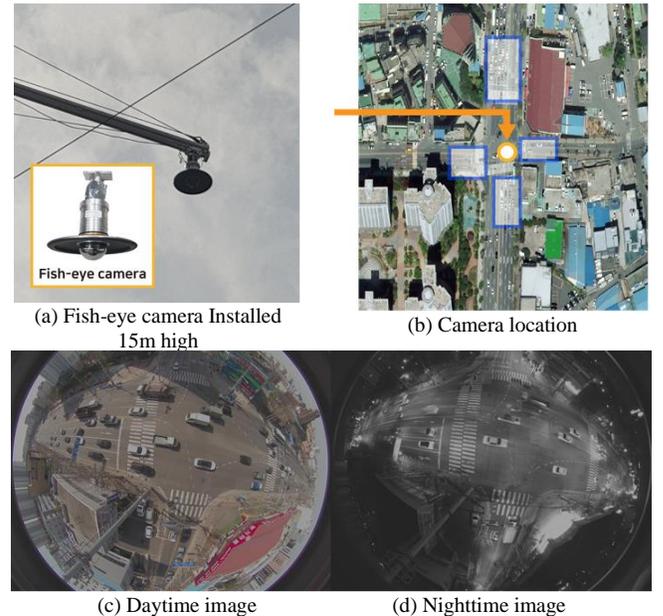

(a) Fish-eye camera Installed 15m high
(b) Camera location
(c) Daytime image
(d) Nighttime image
Fig. 5. Testbed photos.

The testbed was a signalized intersection with 4 legs located in Bucheon city. One thing that differentiates the present experiment from others is the use of a single fish-eye camera that covered all the intersection approaches in a single video frame. This scheme is more economical than other surveillance schemes that require a camera for every intersection approach. Figs. 5 (a) and (b) show the camera installed in the testbed. Figs. 5 (c) and (d) show example photos taken by a fish-eye camera in the testbed. Four images that cover the intersection approaches were cropped and later fed to detectors.

Video was shot for 4 weekdays, and 17,968 images were randomly chosen to train and test the proposed detectors. Half of the images were shot during daytime, and the rest were recorded during nighttime. The testbed images were randomly divided into train, validation, and test sets. The validation and test sets were 10% of the total images, respectively, and thus 80% of the images were used for training. Each image was manually annotated with bounding boxes and ellipses.

To enhance the model performance, backbones of the proposed models were pretrained with an open dataset. The pretrained backbones were then inserted into the proposed

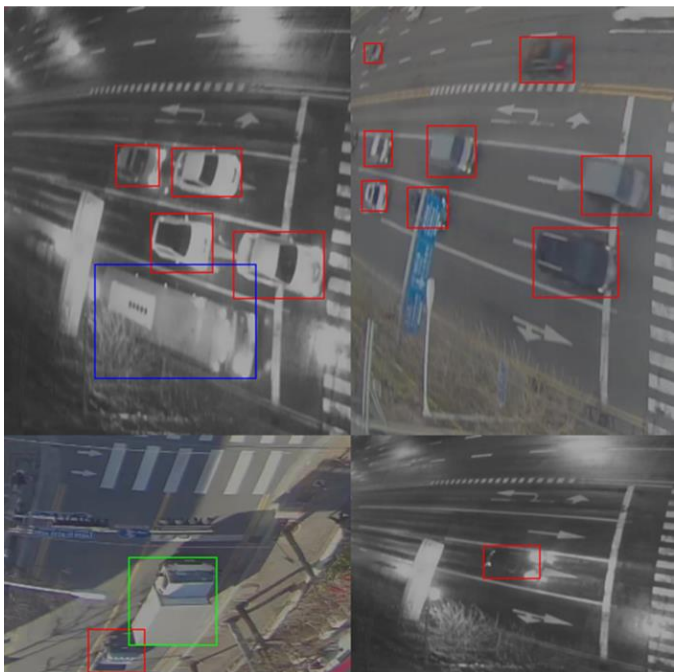
(a) Mosaic example for bounding boxes

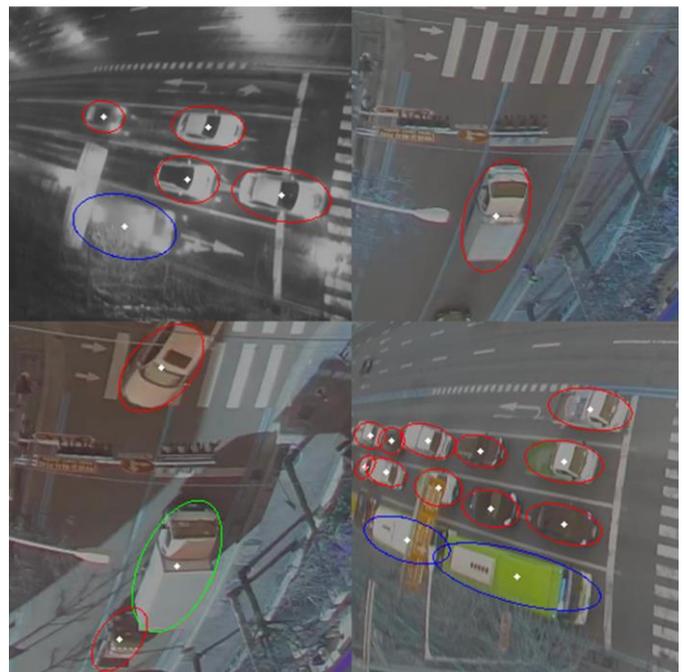
(b) Mosaic example for bounding ellipses

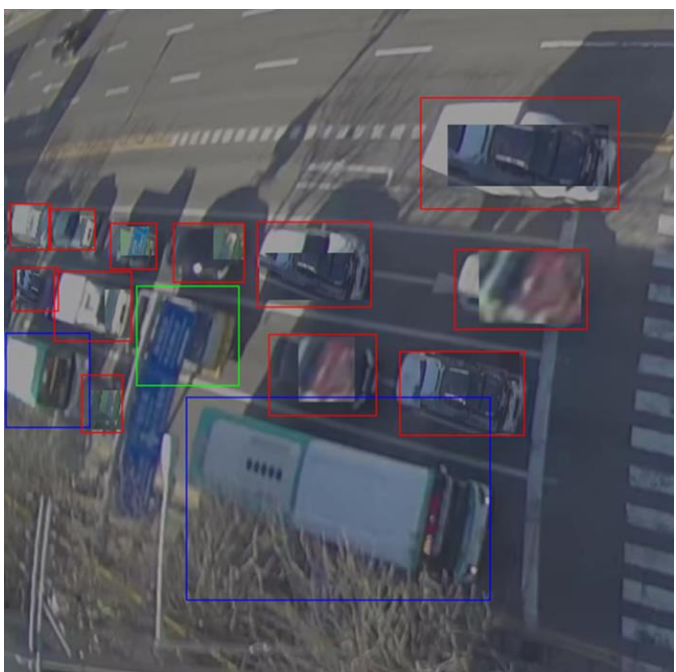
(c) CutMix example for bounding boxes

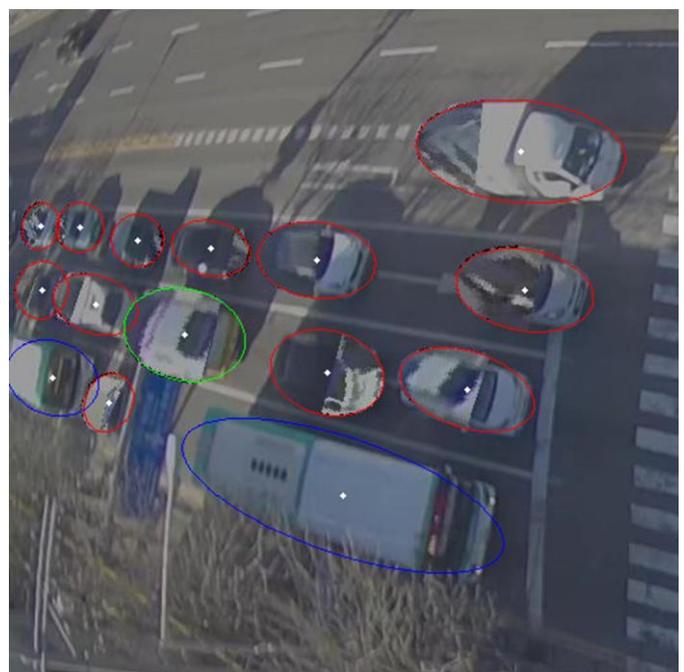
(d) CutMix example for bounding ellipses

Fig. 6. Data augmentation examples.

detection models and fine-tuned on data collected in the testbed. A UA-DETRAC is a representative dataset that was comprised only of vehicle images [52]. The total number of training images in this dataset is tantamount to 80,000.

The current trend to increase detection performance is leaning toward the augmenting of training data. The YOLO version 4 employed various data augmentation schemes and considerably raised the detection accuracy. We also chose three schemes that encompassed label smoothing [53], Mosaic [35], and CutMix [54], each of which proved effective in enhancing the detection accuracy. Three augmentation schemes were independently and collectively tested for both bounding boxes and ellipses. Some mathematical tricks were necessary in order to apply CutMix to the bounding ellipses. Fig. 6 shows typical examples of the images augmented for training models in the present study. We applied MOSAIC and CutMix schemes to the training data off-line, and for each scheme we generated augmented images that equaled the number of original images. The original and augmented images with box and ellipse labels are available in http://00bigdata.cau.ac.kr/.

## V. TESTING THE DETECTION MODELS

The original double-stacked hourglass network (Hourglass-104) was downsized and inserted into both CenterNet and SpotNet to alleviate the computational burden to enhance the vehicle detection performance (see Fig. 1). For reference models, a CSPDarknet53 was used for YOLO4, and A VGG-16 network was used to set up an SSD, because these backbones recorded the best performance in previous works [35],[36]. Basically, all proposed models were trained and tested based on data collected in the testbed. On the other hand, the UA-DETRAC dataset [52], which collects only images for vehicle detectors, was utilized to pretrain the models. Even though pretraining had to be conducted with models based on bounding boxes, CenterNet or SpotNet models with bounding ellipses algorithm was used with a consistent batch size (=8), whereby an Adam optimizer altered the learning rate while training models, and all the proposed models shared a stopping criterion. Each of the proposed models was trained on the same computing environment with a single GPU, which was a NVIDIA Tesla V100 with 32 GB of HBM2 memory.

We evaluated 4 different models based on the same dataset reserved for testing. Both the bounding ellipse and the box was tested for non-anchor-based detectors (CenterNet and SpotNet), and the remaining anchor-based models (YOLO4 and SSD) had to be tested only for bounding boxes. The regression loss was altered according to the types of model specifications that were chosen (see the fourth row of Table 1). A PIOU loss that replaces the conventional regression loss was tested for CenterNet. In order to improve the detection performance, we attempted 3 different augmentation schemes and also applied 3 different combinations of them to the training of each model. For inference with both CenterNet and SpotNet models, a confidence threshold to identify the vehicle presence was set at 0.3 on a trial-and-error basis.

Table 1 lists mAP scores from the test results when models were trained only on local data collected in the testbed. CenterNet with heatmaps augmented by bounding ellipses (mAP = 0.877) outperformed all other detectors with bounding boxes for baseline cases where no data augmentation skill was adopted. This result was achieved owing to adopting a heatmap augmented with bounding ellipses. Heatmaps of the actual shapes of vehicles was very helpful for the regression of the size and orientation of vehicles. As counterevidence, adopting an original heatmap that draws a circle for a vehicle deteriorated the detection performance (mAP=0.785) when accompanied with a regression head that attempted to fit the shape of oriented vehicles. Such an incompatibility was observed even when an original CenterNet used the box regression and adopted original heatmaps with circles (mAP =0.853). Consequently, for

TABLE I
TEST SCORES (MAP) FOR DIFFERENT DETECTION MODELS THAT WERE TRAINED ONLY ON THE TESTBED DATA

| | | CenterNet | | | | | SpotNet | | YOLO v4 | SSD |
|---|---|---|---|---|---|---|---|---|---|---|
| Heatmap | | Original | | | Augmented | | Original | Augmented | N/A | |
| Bounding shape | | Box | Ellipse | OBB | Ellipse | OBB | Box | Ellipse | Box | |
| Regression loss | | W,H | W',H',O | PIOU | W',H',O | PIOU | W,H | W',H',O | W,H,C | |
| | Baseline model | 0.853 | 0.785 | 0.864 | 0.877 | 0.878 | 0.644 | 0.871 | 0.810 | 0.710 |
| | Label smoothing | 0.852 | 0.786 | 0.855 | 0.867 | 0.871 | 0.671 | **0.906** | 0.841 | 0.719 |
| Data | Mosaic | 0.847 | 0.818 | 0.851 | 0.864 | **0.876** | 0.666 | 0.843 | 0.858 | 0.727 |
| augme | CutMix | 0.842 | 0.795 | 0.877 | 0.883 | **0.922** | 0.652 | 0.860 | 0.850 | 0.723 |
| ntation | Mosaic+Label smoothing | 0.885 | 0.786 | 0.807 | **0.912** | 0.889 | 0.612 | 0.838 | 0.840 | 0.727 |
| | CutMix+Label smoothing | 0.860 | 0.857 | 0.901 | 0.947 | 0.929 | 0.752 | 0.911 | 0.880 | 0.721 |
| | Mosaic+CutMix | 0.875 | 0.894 | 0.879 | **0.933** | 0.884 | 0.700 | 0.783 | 0.878 | 0.740 |

C: center point coordinates, H: height of horizontal bounding box, W: width of horizontal bounding box
O: orientation of major axis of bounding ellipse, H': length of major axis of bounding ellipse, W': length of minor axis of bounding ellipse
PIOU: pixels-IOU

could benefit from pretraining because only the backbone was extracted from a pretrained model and used for the next stage of fine-tuning. Our experiment results showed that freeing all weights while fine-tuning the models outperforms fixing the pretrained weights of the backbone.

For a fair comparison, all models shared a training algorithm and hyper-parameters. A stochastic gradient descent (SGD) CenterNet the compatibility between the shape used by a regression head and that by a heatmap turned out to be the most significant aspect in securing detection accuracy. The PIOU loss that depends on OBBs slightly raised the detection performance (mAP=0.878) when accompanied by heatmap augmentation with bounding ellipses. On the other hand, the PIOU loss with augmented heatmaps considerably enhanced the detection performance (mAP=0.878) when compared with the case where the loss was applied to original heatmaps (mAP=0.864). This shows that a heatmap augmented with bounding ellipses also improved the performance of an OBB-based vehicle detection method.

It is meaningful that the performance of non-anchor-based models with augmented heatmaps (CenterNet and SpotNet) was superior to that of anchor-based models (YOLO v4 and SSD) which are mainstream learning models in the object detection.

detection performance (mAP=0.924). The overall ranking among baseline models with pretraining was similar to that among the baseline models without pretraining. Also, as in the previous case without pretraining, when a PIOU loss based on OBBs was adopted, heatmaps augmented with bounding ellipses considerably increased the mAP score of CenterNet (=0.924), when compared with the score (=0.865) of the PIOU loss using the original heatmaps.

Regarding data augmentation schemes applied to the

TABLE II
TEST SCORES (MAP) FOR DIFFERENT DETECTION MODELS THAT WERE PRETRAINED ON THE UA-DETRAC DATASET

|  |  | CenterNet |  |  |  |  | SpotNet |  | YOLO v4 | SSD |
|---|---|---|---|---|---|---|---|---|---|---|
|  | Heatmap | Original |  |  | Augmented |  | Original | Augmented | N/A |  |
|  | Bounding shape | Box | Ellipse | OBB | Ellipse | OBB | Box | Ellipse | Box |  |
|  | Regression loss | W,H | W',H',O | PIOU | W',H',O | PIOU | W,H | W',H',O | W,H,C |  |
| Data augmentation | Baseline model | 0.873 | 0.848 | 0.865 | 0.917 | 0.924 | 0.721 | 0.897 | 0.873 | 0.819 |
|  | Label smoothing | 0.866 | 0.853 | 0.887 | 0.953 | 0.916 | 0.741 | 0.924 | 0.878 | 0.821 |
|  | Mosaic | 0.836 | 0.850 | 0.864 | 0.868 | 0.869 | 0.678 | 0.862 | 0.862 | 0.778 |
|  | CutMix | 0.879 | 0.859 | 0.868 | 0.933 | 0.936 | 0.725 | 0.924 | 0.896 | 0.776 |
|  | Mosaic+Label smoothing | 0.874 | 0.814 | 0.847 | 0.916 | 0.853 | 0.650 | 0.860 | 0.800 | 0.802 |
|  | CutMix+Label smoothing | 0.884 | 0.892 | 0.896 | 0.949 | 0.930 | 0.836 | 0.936 | 0.919 | 0.756 |
|  | Mosaic+CutMix | 0.870 | 0.923 | 0.803 | 0.937 | 0.854 | 0.717 | 0.826 | 0.902 | 0.819 |

C: center point coordinates, H: height of horizontal bounding box, W: width of horizontal bounding box
O: orientation of major axis of bounding ellipse, H': length of major axis of bounding ellipse, W': length of minor axis of bounding ellipse
PIOU: pixels-IOU

A SpotNet was employed to raise the detection performance under the assumption that matching the actual scale in a segmentation map could resolve the problem of having to perform regression fitting with respect to a downsized heatmap. The detection performance of SpotNet (mAP=0.871) was, however, inferior to that of CenterNet (mAP=0.877), although an additional semantic segmentation head was added. This could have been the result of inaccuracies in drawing the ground-truth segmentation map. The role of a segmentation map drawn with bounding ellipses was overlapped with that of heatmaps. If a finer segmentation map could be manually drawn, the detection performance would have been enhanced.

The contribution of data augmentation turned out to be salient in improving the detection accuracy. The top mAP score (=0.947) was recorded when applying the combination of a CutMix and label-smoothing technics to CenterNet with augmented heatmaps. This score is very exceptional in vehicle detection for traffic surveillance. Overall, the simple adjustment of drawing bounding ellipses on the ground-truth heatmap together with cutting-edge data augmentation tools significantly raised vehicle detection capability.

Table 2 shows the test results when each model was fine-tuned on local data after being pretrained on a UA-DETRAC dataset. The data augmentation schemes were applied only to the local data for fine-tuning rather than to the UA-DETRAC data. For baseline cases where the pretrained models were fine-tuned without data augmentation, the detection performance of all models was better than that without pretraining.

Among baseline cases of pretrained models, CenterNet with heatmaps augmented with bounding ellipses also recorded a better performance (mAP=0.917) than those using bounding boxes (mAP=0.873), as in the baseline cases without pretraining. Similarly, the PIOU loss slightly ameliorated the pretrained models, the label smoothing scheme led to the greatest improvement for CenterNet detection with bounding ellipses (mAP=0.953). The mAP score exceeded 95%, which has rarely been achieved in vehicle detection studies for traffic surveillance. A CenterNet approach with the combination of CutMix and label smoothing techniques recorded the second-best performance when bounding ellipses were used (mAP=0.949). According to the experimental results of Chen. et al. [27], among object detection methods based on OBBs, the PIOU loss method recorded the highest accuracy. It should be, however, noted that the best mAP (=0.953) of the proposed approach exceeded that of the PIOU loss method (=0.924).

TABLE III
TEST PERFORMANCE (MAPs) ACCORDING TO THE NUMBER OF ANNOTATED IMAGES FOR FINE-TUNING

| The number of annotated images for fine-tuning | CenterNet | SpotNet |
|---|---|---|
| 14,372(100%) | 87.7% | 87.1% |
| 8,623(60%) | 84.2% | 83.5% |
| 7,186(50%) | 80.0% | 80.7% |
| 5,749(40%) | 78.4% | 77.0% |
| 4,311(30%) | 75.4% | 75.7% |

No combination of data augmentation schemes could produce a YOLO v4 that was superior to the best non-anchor-based detection approach with augmented heatmaps. This verifies that non-anchor-based detection approaches using bounding ellipses outperformed "state-of-the-art" anchor-based models for vehicle detection. The superiority comes from the possibility that vehicles can be delineated using ellipses without margins. The superiority of the proposed method cannot be generalized for detecting various objects in a more complex shape.

Even though the present study did not invent a new detection architecture of neural net, our main contribution was obtaining such a cenotaphic performance in detecting vehicles by finding the best combination of existing cutting-edge technologies. Of course, several modifications were suggested to adapt CenterNet for adequate vehicle detection.

Even though a pretrained CenterNet had a good performance for vehicle detection, a large local dataset including more than 17,968 annotated images was used for fine-tuning, validating, and testing. This required a great deal of human effort, and manually drawing the bounding ellipses for every site for traffic surveillance would not be sustainable. To reduce human effort, we conducted sensitivity analysis to identify how much local data are necessary to secure an acceptable level of accuracy. The relationship between the number of images used for fine-tuning and the test accuracy is shown in Table 3. Fine-tuning pretrained CenterNet and SpotNet models using about 8,600 annotated images acquired a performance that almost matched that of the models fine-tuned on the total number of training images. When only 30% of images were used for fine-tuning, a mAP score higher than 75.4% was obtained using CenterNet.

In the present study we conducted an experiment to quantify the inference speed of models. The inference time was measured according to the number of video frames that could be processed within a second. The average number of frames per second (FPS) is shown in Table 4. The experiment was conducted in a computing environment with a single GPU, which was a NVIDIA Tesla V100 with 32 GB of HBM2 memory. Surprisingly, CenterNet recorded a faster detection speed than the YOLO v4. CenterNet and SpotNet had an advantage in speed that was obtained by a lighter backbone. As mentioned earlier in section III, the original Hourglass-104 was considerably downsized without a loss of detection performance, which is a significant contribution of the present study. CenterNet and SpotNet processed 38 and 31 FPS, respectively. Such speeds are sufficient to track vehicles for use in traffic surveillance, because traffic surveillance has no need to track hundreds of video frames per second.

Fig. 6 shows examples of success and failure in vehicle detection using CenterNet and SpotNet models. In successful cases, the two non-anchor-based detectors can identify vehicles with bounding ellipses that have no margins. The orientation of vehicles is also accurately determined. Most failures occurred when vehicles were doubly detected. Promising results are reaped when there are no false positives or missing vehicles. Both models also capably detected vehicles at nighttime.

VI. CONCLUSION

The present study is an investigation into the possibility of enhancing the performance of vehicle detection by using a bounding ellipse instead of a bounding box. The conventional object detection approaches are based on anchor boxes and cannot be used with bounding ellipses, because they require the computations of NMS and IOU, which increases the computing complexity when bounding ellipses are used. Two non-anchor-based object detectors were chosen to test the utility of bounding ellipses.

TABLE IV
INFERENCE SPEED OF DETECTION MODELS

| Model | Bounding | Average frames per second (FPS) |
|---|---|---|
| CenterNet | Ellipses | 38 |
| | Boxes | 38 |
| | OBB(PIOU) | 38 |
| SpotNet | Ellipses | 31 |
| | Boxes | 31 |
| YOLO4 | Boxes | 35 |
| SSD | Boxes | 24 |

Using bounding ellipses considerably enhanced the performance of vehicle detection. In particular, pretraining models with an open UA-DETRAC dataset composed only of vehicle images considerably improved the detection accuracy. A CenterNet model that was fine-tuned on heatmaps augmented by bounding ellipses recorded the highest mAP score when labels were smoothed. We obtained good accuracy even when a smaller number of annotated images were available for fine-tuning once a robust backbone was pretrained on a large open dataset. Such transfer learning makes it possible to deploy a detection model to area-wide traffic surveillance with minimal human effort.

It should be noted that the proposed detection scheme with bounding ellipses was verified only for vehicle detection at a single site. The proposed approach must be tested in the future at many other sites. It will be a challenge to secure images annotated with bounding ellipses for each site, even though we minimized the amount of data required. However, other models with bounding boxes would have the same burden for labeling.

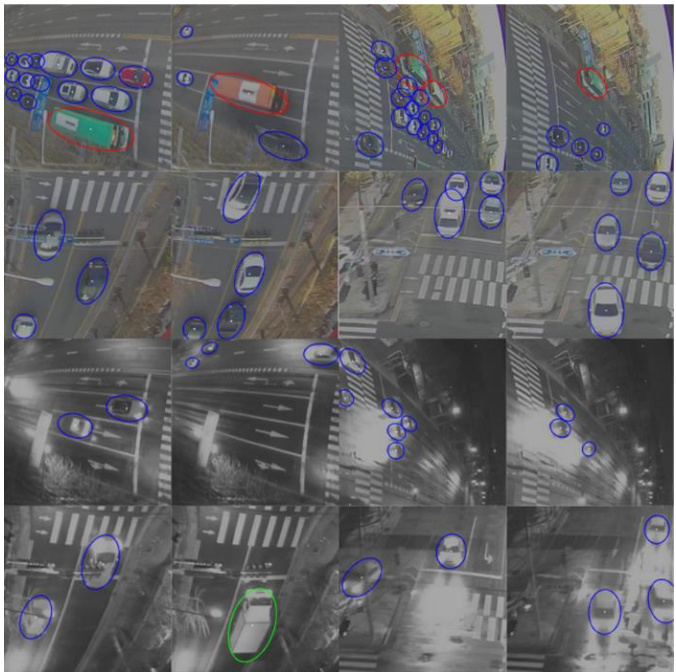
(a) Successful examples of CenterNet

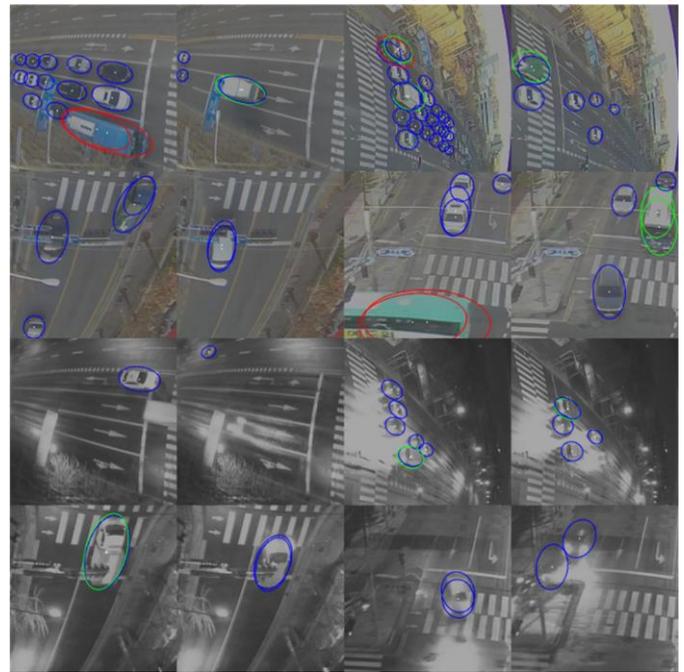
(b) Failed examples of CenterNet

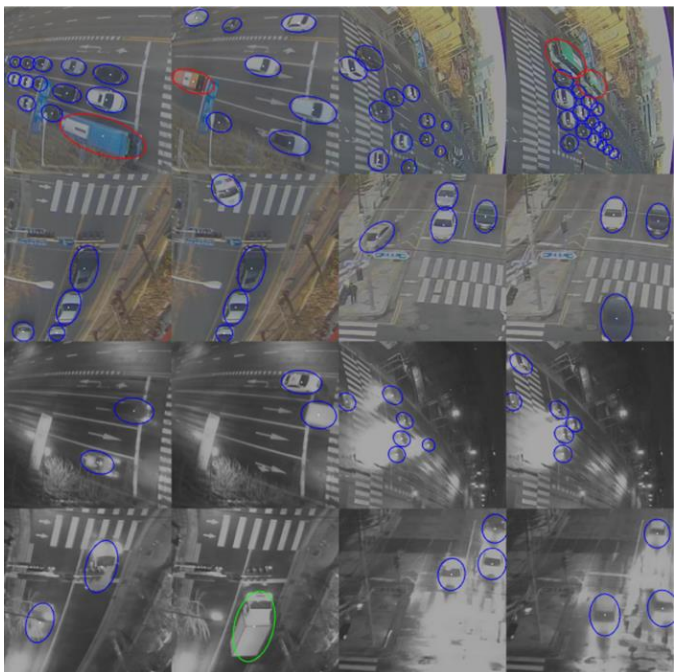
(c) Successful examples of SpotNet

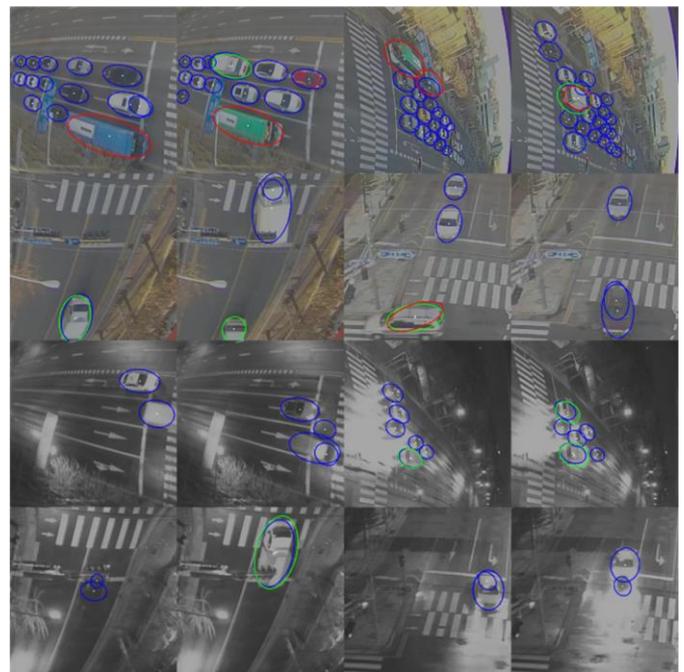
(d) Failed examples of SpotNet

Fig. 6. Examples of success and failure in vehicle detection.

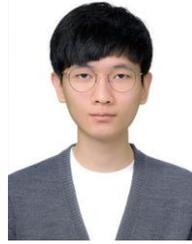

**Byeonghyeop Yu** is with the Laboratory of Big-data Applications in Public Sector, Chung-Ang University as a graduate student. He is preparing his thesis for an M.S. degree. The thesis theme covers applying an attention-based graph neural network to forecasting the road speed and travel time.

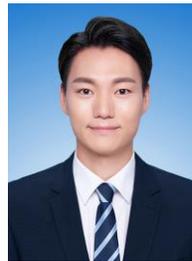

**Johyun Shin** is with the Laboratory of Big-data Applications in Public Sector, Chung-Ang University. He is preparing his thesis for an M.S. degree. The thesis theme is to measure traffic parameters such as volume, speed, and density based on deep learning.

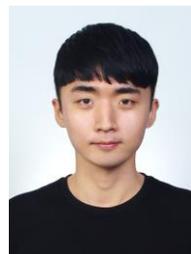

**Gyeongjun Kim** is with the Laboratory of Big-data Applications in Public Sector, Chung-Ang University. He is a graduate student preparing his thesis for an M.S. degree. The thesis theme is traffic light control based on reinforcement learning

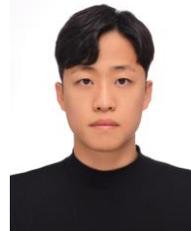

**Seungbin Roh** is with the Laboratory of Big-data Applications in Public Sector, Chung-Ang University. He is preparing his thesis for an M.S. degree, the theme of which covers applying an autoencoder to measuring traffic volumes without manually tagging images.

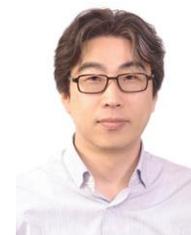

**Keemin** Sohn is a Professor with the Department of Urban Engineering, Chung-Ang University. His research interests include applications of artificial intelligence to transportation engineering and planning.